\documentclass[11pt,a4paper]{article}
\usepackage[hyperref]{emnlp2020}
\usepackage{times}
\usepackage{latexsym}
\usepackage{graphicx}
\usepackage{amsfonts}
\usepackage{booktabs}
\usepackage[full,warn]{textcomp}
\newcommand{\qdist}[1]{\ifmmode\langle#1\rangle\else\textlangle#1\textrangle\fi}

\newcommand{\eric}[1]{{\textcolor{black}{#1}}}

\usepackage{microtype}

\aclfinalcopy 

\title{E-commerce Query-based Generation based on User Review}

\author{Yiren Liu \\
  University of Illinois \\
  \texttt{yirenl2@illinois.edu} \\
  \And
  Kuan-Ying Lee \\
  University of Illinois \\
  \texttt{kylee5@illinois.edu} \\
}

\date{}

\begin{document}
\maketitle

\begin{abstract}
With the increasing number of merchandise on e-commerce platforms, users tend to refer to reviews of other shoppers to decide which product they should buy. However, with so many reviews of a product, users often have to spend lots of time browsing through reviews talking about product attributes they do not care about. We want to establish a system that can automatically summarize and answer user's product specific questions.

In this study, we propose a novel seq2seq based text generation model to generate answers to user's question based on reviews posted by previous users. Given a user question and/or target sentiment polarity, we extract aspects of interest and generate an answer that summarizes previous relevant user reviews. Specifically, our model performs attention between input reviews and target aspects during encoding and is conditioned on both review rating and input context during decoding. We also incorporate a pre-trained auxiliary rating classifier to improve model performance and accelerate convergence during training. Experiments using real-world e-commerce dataset show that our model achieves improvement in performance compared to previously introduced models.

\end{abstract}

\section{Introduction}
Most online e-commerce platforms display user generated reviews containing information that might be helpful for users' decision making. These reviews usually come with great quantity and are posted in informal tone of speech. This makes it very hard for common users to scan through all the reviews for some particular information they need. By doing this study, we want to propose a model to extract useful information from both item description and user generated reviews that can help users during decision making and generate explanations containing both text and images that are easy to read and understand for the target user.\\
Many works attempt to approach this problem by generating reviews using NLG models trained on question-answer pairs. Previous study \cite{chen_review-driven_2019} has introduced a seq2seq based generation method that generates answer to question conditioned on auxiliary review snippets extracted from reviews under the target item. The method exhibits promising performances over real-world e-commerce dataset. However, the underlying premise of these QA pair based methods requires sufficient amount of existing QA pairs under each applied categories for training purpose. For some cases where exists plenty of reviews, still the amount of QA pairs enough for model training might be luxurious in practice. \\
Some other recent studies focus on directly generating review-like explanation based only on existing reviews. Ni et al. \cite{ni_justifying_2019} introduce a method in generating recommendation justification using user review based on fine-grained aspects. This model takes as input both reviews under the specific item and historical reviews generated by the target user, and synthesizes justification in a review-like semantic format. The paper utilizes the aspect-planning generation technique to ensure that the generated justification contains certain aspects or attributes beforehand. Although this method effectively increases the diversity and relevance of the generated justification, it might cause manipulation of semantic meaning from the original input and negatively impact the factual accuracy of the generated justification.\\

\begin{figure*}
\centering
\includegraphics[width=\textwidth]{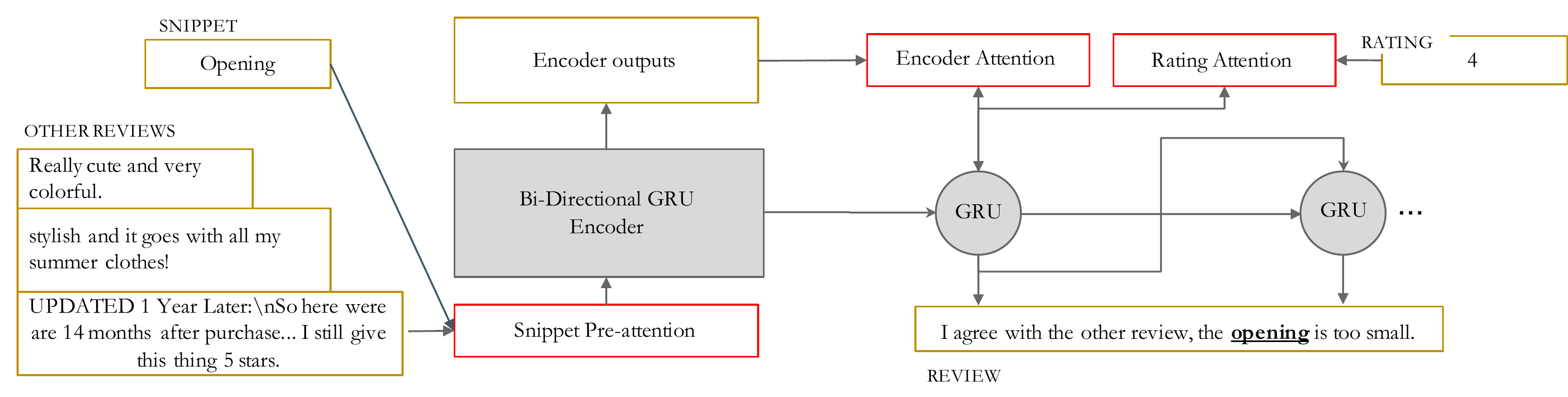}
\caption{Based on other reviews of the same item and given aspects (snippets) extracted from the input question (testing) or target review (training), our model learns to generate a question-aware answer. Our framework also takes into consideration review rating to accelerate convergence in training and generate reviews with specific sentiment polarity per user preference in testing.}
\end{figure*}

In this work, we introduces a model that generates attribute-focused review summary as answer to users' item-specific question. To increase the quality of the generated answer, we extract text snippets describing item attributes from reviews and perform conditional NLG using our proposed model. Additionally, we take into account the sentimental polarity of each review to increase the factual consistency between the original reviews and the synthesized summary.
The main contributions of this paper are as follows:
\begin{itemize}
\item We propose a model utilizing attention and copy mechanism to perform jointly-conditional natural language generation based on both item attribute and rating information from user review. 
\item We show the probability of adopting ACGAN-like discriminator structure to help improve the rating-conditioned generation performance based on our model design.
\end{itemize}



\section{Framework}
Our target is to generate an answer to a question the user poses about a certain product by referring to the existing reviews of that product. Specifically, our framework has three inputs. Existing reviews, a question from the user and \eric{a rating specified by the user that decides the reference reviews. The rating guides the selection of reference reviews and help the user see the commonality amongst compliments or criticism. (eg. a specification of five demonstrates what five-star reviews have in common)}

However, the question answer pairs could be absent for many recent products or the platform only provides review section and not the QA section. Under this circumstance, the next question is then how we could train a model that takes in a question and answer accordingly without QA pairs available?

Without question answer pair, we propose to simulate each review as a question. That is, we would like to extract snippets from it and use these snippets as aspects of interest. Note that during testing, we could simply run the same snippet extraction procedure and obtain aspects of interest from the question. Doing so, we could get away with the absence of QA pair. Also, we unlike many existing work \cite{gao_product-aware_2019, yu_review-aware_2018,kulkarni_productqna_2019}, we do not have to pre-define aspects of interest and instead let the model learns directly from the reviews. 

In training, our model first encodes all but the target review, and reconstruct the review by conditioning on the input snippets and the rating of the same review, while in testing, the model encodes all the reviews and generates the answer based on the extracted snippets from the question (and the specified rating).

\section{Approach}
We first introduce the problem formulation as follows. Given all other item reviews $X_{-i} = (X_1, ... X_{i-1}, X_{i+1}, ... X_{l_s})$, where $X_i$ denotes the $i_{th}$ review, snippets $S = (S_1, ... S_n)$ where $S_j$ denotes the $j_{th}$ snippet and rating $R$, our goal is to maximize the conditional probability $p(X_i | X_{-i}, S, R)$. The conditional probability could be expressed as:
\begin{equation}
p(X_{i} | X_{-i},S,R) = \prod_{j=1}^{l_r}{ p(X_{i,t} | X_{i,j<t},X_{-i},S,R) }
\end{equation}
where $X_{i,j<t} = (X_{i,1}, ... X_{i,t-1})$ and $l_r$ denotes the length of the target review.

Our model could be divided into three parts, (1) encoder that encodes the existing reviews, (2) decoder that generates the answer and (3) discriminator that judges the rating and the factuality of the generated answer. Our model is trained upon \eric{three losses, the cross entropy loss between the generated and the groundtruth words, the GAN loss from the discriminator and the classification cross entropy loss by the auxiliary rating classifier.}

\subsection{Item Review Attention Encoder} 
In order to obtain context vector of the model input, we use three separate encoders for item reviews, review attribute snippets and review ratings.
For item reviews $\mathbf{X}$, we first obtain the embedding vector of the input and pass them through a vanilla bi-directional GRU to get the last layer hidden states as the context vector $\mathbf{H} \in \mathbb{R} ^{l_{r}, l_{s},l_{emb}}$:
\begin{equation}
\mathbf{X}_{emb} = embedding(\mathbf{X})
\end{equation}
\begin{equation}
\mathbf{H} = \overrightarrow{GRU(\mathbf{X}_{emb})} + \overleftarrow{GRU(\mathbf{X}_{emb})}
\end{equation}
Where $\mathbf{X}_{emb} \in \mathbb{R} ^{l_{r}, l_{s}}$ stands for the embedding vectors for reviews under the specified item, $l_{r}$ and $l_{s}$ stand for the number of reviews and the length of each review. 
Similarly for attribute snippet and review rating input, we obtain context vector $\mathbf{S}$ and $\mathbf{U}$. Since attribute snippet and rating contains no positional information, we only obtain their embeddings as their context vector.

\textbf{Snippet-conditioned Generation.}
Based on the bi-directional attention method introduced in Seo et al.'s work\cite{seo_bidirectional_2018}, we use a bi-directional attention mechanism between item review context vector and snippet context vector to obtain snippet-conditioned review context vector:
\begin{equation}
\mathbf{A}_{i,j}^{H} = \alpha_{1}(\mathbf{H}_{i},\mathbf{S}_{j})
\end{equation}
\begin{equation}
\mathbf{a}_{i}^{H} = softmax(\mathbf{A}_{i,:}); \mathbf{a}_{j}^{S} = softmax(\mathbf{A}_{:,j})
\end{equation}
Where $\mathbf{H}_{i}$ and $\mathbf{S}_{j}$ represent the $i$ word in item review and $j$ word in attribute snippet and $\mathbf{A}_{i,j}^{H}$ is the attention weight matrix between review word $i$ and snippet word $j$ calculated using attention method $\alpha_{1}$. In this paper, we use $\alpha_{1}(x,y) = v_{\alpha_{1}}^{T}[x,y,x\odot y]$, where $v_{\alpha_{1}}$ is a trainable projection vector. By applying softmax normalization, we get attention scores between review word $i$ and each word from the snippets denoted as $\mathbf{a}_{i}^{H}$, and $\mathbf{a}_{j}^{S}$ vice versa.
After this, we compute the weighted context vector by aggregating the dot product of the attention score and input context vector of both review and snippets:
\begin{equation}
\tilde{\mathbf{H}}_{i} = \sum_{j}{\mathbf{a}_{i,j}^{H}\mathbf{H}_{i}};  \tilde{\mathbf{S}}_{j} = \sum_{i}{\mathbf{a}_{j,i}^{S}\mathbf{S}_{j}}
\end{equation}
\begin{equation}
\tilde{\mathbf{H}}_{S} =  \alpha_{2}(\mathbf{H},\tilde{\mathbf{H}},\tilde{\mathbf{S}}) \odot \mathbf{H}
\end{equation}
Here $\mathbf{a}_{i,j}^{H}$ denotes the $j$ element in $\mathbf{a}_{i}^{H}$ and  $\mathbf{a}_{j,i}^{H}$ denotes the $i$ element in $\mathbf{a}_{j}^{H}$. $\tilde{\mathbf{H}}_{S}$ denotes the weighted context for item review text, and similarly to equation 1 $ \alpha_{2}$ denotes the attention method. Here we use $\alpha_{2}(x,y,z) = v_{\alpha_{2}}^{T}[x;y;x\odot y;x\odot z]$ to emphasize on the original input context.

\textbf{Rating-conditioned Generation.}
To enforce rating condition over generation, we perform attention weighting between query input rating embedding $\mathbf{r}$ and each input item review rating embedding $\mathbf{R}$:
\begin{equation}
\tilde{\mathbf{R}}_{t} = \beta_{1}(\mathbf{r},\mathbf{R}_{t}) \odot \mathbf{R}_{t}
\end{equation}
Where $\tilde{\mathbf{R}}_{t}$ denotes weighted rating context for item review $t$, and we use attention method $\beta_{1}(x,y) = v_{\beta_{1}}^{T}(x \odot y)$ to emphasize the rating similarity between reviews.

\subsection{Decoder}
We use bi-directional GRU for decoding. To ensure the generated output is relevant to the input, we calcualte attention using the context vector we obtained above and the decoder hidden states $h$ and project the representation onto the vocabulary dimension: 
\begin{equation}
\mathbf{a}_{t}^{H} = softmax(\mathbf{W}_{H}^{T}[\tilde{\mathbf{H}}_{t};h] + \mathbf{b}_{H})
\end{equation}
\begin{equation}
\mathbf{a}_{t}^{R} = softmax(\mathbf{W}_{R}^{T}[\tilde{\mathbf{R}}_{t};h] + \mathbf{b}_{R})
\end{equation}
\begin{equation}
P(w_{t}) = softmax(\mathbf{W}^{T}[h;\mathbf{a}_{t}^{H}\odot \tilde{\mathbf{H}}_{t};\mathbf{a}_{t}^{R}\odot \tilde{\mathbf{R}}_{t}] + \mathbf{b})
\end{equation}
Where the $P(w_{t})$ denotes the probability of word $w_{t}$ appearing at this decoding step, and $\mathbf{W}$ and $\mathbf{b}$ denote the representation of a dense layer.

We use negative log likelihood as the generation loss:
\begin{equation}
L_{gen}(\theta_g) = -\sum_{i}^{N}w_{t}log\hat{w_{t}}
\end{equation}
Where $L_{gen}(\theta_g)$ is the loss for model parameters $\theta_g$ at each decoding step.

\begin{table*}[t]
\centering
\begin{tabular}{@{}l|llllll@{}}
\toprule
{} & \textit{BLEU-1} & \textit{BLEU-2} & \textit{BLEU-3} & \textit{BLEU-4} & \textit{METEOR}& \textit{ROUGE-L}    \\ 
\midrule
Random & \textbf{13.279}           & 4.652          & 1.693              & 0.651               &6.248              & 8.341              \\
NN-rating & 13.022               & 4.501         & 1.563                & 0.592                &6.460               & 8.639               \\ 
Seq2seq & 8.932               & 5.543          & 3.466               & 2.227               & 7.452               & 23.134              \\ 
Dong et al. &0.616               &0.182        &0.054                & 0.012                & 1.652              & 6.919               \\ 
Ni et al. & 8.026              & 5.192          & 3.389                & 2.278                & 7.861               & 25.529             \\ 
\midrule
Ours  & 8.920              &5.699         & 3.670               & \textbf{2.455}                & \textbf{8.767}               &\textbf{26.939}                          \\ 
Ours w/o rating  & 9.359              &\textbf{5.910}         & \textbf{3.753}               & 2.448                & 8.140               & 24.484  \\ 
\bottomrule
\end{tabular}
\caption{
\label{table 1} Comparison of Model Performance based on Evaluation Metrics
}
\end{table*}

\subsection{Rating Classifier}
To further ensure that the generated answers is consistent to the input rating, we employ an auxiliary classifier that takes in the generated answer and computes the cross entropy loss between the predicted and the groundtruth rating. To ensure that the classifier is strong enough, we pre-train it on all training sentences and formulate the target as a sentiment analysis task.

And to handle cases when user simply wants a overall review by not specifying the input rating, we add an additional \qdist{PAD} token during training that signals the generator to generate an overall review. In these circumstances, the classification cross entropy loss will not be computed and optimized (since the classifier only has ratings from one to five but not the \qdist{PAD} token).

The classification loss is
\begin{equation}
L_{cls}(\theta_c) = -\sum_{c=1}^{5}{y_c \log{p_c}}
\end{equation}
where $y_c$ is the groundtruth probability (either 0 or 1) for rating $c$, $p_c$ is the prediction probability for rating $c$, and $\theta_c$ is the classifier parameters at each timestamp. 

We combine the classification loss with the generation loss as the final loss function of our model:
\begin{equation}
L(\theta_g,\theta_c) = -\lambda \sum_{i}^{N}w_{t}log\hat{w_{t}} -  (1 - \lambda) \sum_{c=1}^{5}{y_c \log{p_c}}
\end{equation}
Where $\lambda \in [0,1]$ is a hyper-parameter.





\section{Experiments}

\subsection{Dataset}
We use the Amazon Review Data\footnote{\url{https://nijianmo.github.io/amazon/index.html}} to train and validate our model performance in review generation. We use data under Amazon Fashion subcategory and  filter out items with less than 20 reviews to ensure robustness of data. 
For review attribute snippet extraction, we use the noun phrase extraction method introduced by Pan et al. \cite{pan_cross-lingual_2017}. 

\subsection{Implementation Detail}
Our model is implemented using Pytorch\footnote{\url{https://www.pytorch.org}}. Data tokenization is conducted using NLTK\footnote{\url{https://www.nltk.org}}. We set the hidden size of both the encoder and decoder to 512 and layer number to 1. For optimizer we use Adam with a learning rate of 0.0002. The embedding size is set to 512. We also apply a gradient clipping of $[-5,5]$ to prevent exploding gradient. The maximum input review number is set to 20 and the maximum length of review and attribute snippet is also set to 20. We truncate the review and words exceeding the maximum limit. We use beam search with a beam size of 5 during decoding, and the decoding maximum length is set to 15.

\subsection{Baseline}
We compare with several baselines and previous work of similar target.

\textbf{Random.} It randomly selects one review from all reviews of the target product.

\textbf{NN-rating.} A Nearest Neighbor based method that first selects all reviews of the query product. Then, if user provides a target rating, it randomly selects one of review that has the same rating. Otherwise, it randomly selects one amongst all of them.

\textbf{Seq2seq.} A sequence-to-sequence model with attention mechanism \cite{luong2015effective}. This model, as ours, takes input all reviews but encodes them without snippet and rating information. The decoder attends to the rating, the snippets and the encoder outputs to produce context vector during decode.

\textbf{Dong.} Model by Dong et al. \cite{dong-etal-2017-learning-generate}. The model takes input the productID and rating (not the userID as in the original paper due to a user might not be in the training set). The model is then trained to generate review as similar as possible to the original review.

\textbf{Ni.} Reference-based model with aspect-planning proposed by Ni et al. \cite{ni_justifying_2019} on recommendation justification that predicts what review a specific user would give to a specific item. It is based on historic reviews of the user, reviews of the item and fine-grained aspects of the item. We replace the fine-grained aspects input by the extracted snippets. Again, since we target at QA system and has no access to historic user review, we remove the historic user review input module. Hence, the final input would be the snippets along with the item reviews.  


\subsection{Experimental Result}
For evaluation, we use BLEU\cite{papineni_bleu_2002}, METEOR\cite{banerjee_meteor_2005} and ROUGE-L\cite{lin_rouge_2004} scores as our evaluation metrics.

As shown in table \ref{table 1}, our model achieves the best performance on BLEU-4, METEOR and ROUGE-L scores. This shows an improvement in both diversity and precision of generation using our proposed method. We also observe a drop of performance by our model in lower dimension phrase matching based criterion including BLEU-1 to 3. This might be explained as a compromise in establishing a language model with higher diversity, which can be further testified with the higher METEOR and ROUGE-L score.

\section{Related Work}

\noindent\textbf{E-commerce Question Answering} There has been some previous studies focusing on automatic user question answering on E-commerce platforms. \cite{zhang_explicit_2014} introduced a lexicon based method to extract review feature-opinion pairs. These pairs are later utilized in a modified matrix factorization recommendation model to learn the latent representation of user and item based on extracted features and opinions. This makes the recommendation result explainable regarding item features and can be used for item specific question answering. \cite{yu_review-aware_2018} introduced a auto-encoder based model to utilize review information in generating yes-no answer to users' questions. This method could only be used to settle a limited amount of questions in practice. To improve the diversity of generated answers, studies also employ text generation methodology to directly generate answers and explanations. \cite{gao_product-aware_2019} proposed a seq2seq based model with discriminator to incorporate both review information and annotated attribute-value pairs to generate answers. \cite{ni_justifying_2019} introduced a seq2seq model and a masked language model with aspect-planning technique during generation, using item aspect extracted with semi-supervised method. These methods require either fully or partially annotated review item attribute information. We use a fully automatic attribute information extraction method in our proposed model. Beside review and attribute information, we also incorporate review rating data as a condition during generation. 

\noindent\textbf{Seq2seq Generation Framework}
Seq2seq based models are one of the most frequently used models among recent text generation and neural machine translation research. \cite{Mikolov_Recurrent_2010} proposed language model learning using recurrent neural network. To incorporate more inter-phrases correlation during generation, \cite{luong_effective_2015} and \cite{bahdanau_neural_2016} utilized attention mechanism to effectively improve the performance of text generation.

\section{Conclusion}
In this paper, we investigated the question of generating answer to user's item specific question based on review and rating information on e-commerce platform. We proposed a seq2seq based model with attention mechanism to perform conditional generation on both review attributes and ratings. The experiment results showed that our model achieved a better general performance than baseline methods. We also showed that by explicitly incorporating both review attribute information and review rating using attention mechanism and auxiliary rating classifier, the performance of the generation model could be improved.

\bibliography{emnlp2020}
\bibliographystyle{acl_natbib}

\end{document}